\documentclass[runningheads]{llncs}
\usepackage[hidelinks]{hyperref}
\usepackage{subcaption}

\usepackage{makeidx}
\usepackage{graphicx}
\usepackage{amsmath,amssymb} % define this before the line numbering.
\usepackage{color}
\newcommand{\method}[1]{{\textsc{#1}}}
\graphicspath{{figures/}}

\begin{document}
\pagestyle{headings}
\mainmatter

% Insert your submission number here
\def\GCPR19SubNumber{73}

% Replace with your title
\title{Visual Coin-Tracking: Tracking of Planar Double-Sided Objects}

% DO NOT MODIFY these for the draft version that is used for the
% review process.
\titlerunning{Visual Coin-Tracking: Tracking of Planar Double-Sided Objects}
\authorrunning{Jon\'{a}\v{s} \v{S}er\'{y}ch \and Ji\v{r}\'{i} Matas}
\author{
  Jon\'{a}\v{s} \v{S}er\'{y}ch \and
  Ji\v{r}\'{i} Matas}
\institute{CMP Visual Recognition Group, Dept. of Cybernetics, Faculty of Electrical Engineering, Czech Technical University in Prague, Czech Republic}

\maketitle

\begin{abstract}
We introduce a new video analysis problem -- tracking of rigid planar objects in sequences where both their sides are visible. Such \emph{coin-like objects} often rotate fast with respect to an arbitrary axis producing
unique challenges, such as fast incident light and aspect ratio change and rotational motion blur.
Despite being common, neither tracking sequences containing coin-like objects nor suitable algorithm have been published.

As a second contribution, we present a novel \emph{coin-tracking benchmark} containing 17 video sequences annotated with object segmentation masks.
Experiments show that the sequences differ significantly  from the ones encountered in standard tracking datasets.
We propose a baseline coin-tracking method based on convolutional neural network segmentation and explicit pose modeling. Its performance confirms that coin-tracking is an open and challenging problem.
\end{abstract}

\section{Introduction}
Visual tracking is one of the fundamental problems in the field of computer vision.
Given a video sequence and some defined object, e.g. by its location in the first frame, the task is to find its pose in each frame of the sequence. Until recently, standard visual tracking datasets like~\cite{VOT_TPAMI} or~\cite{wu15otb} have been only annotated using bounding boxes and subsequently, state-of-the-art trackers usually represented the objects pose as a rotated or axis-aligned bounding box.
Recently, tracking-by-segmentation, also called video object segmentation, has gained on popularity, thanks to the introduction of segmentation-annotated datasets like DAVIS~\cite{perazzi2016davis} and YouTube-VOS~\cite{xu2018youtube}. Here, object pose is a segmentation mask.

Visual tracking is an active research field; tracker performance improves significantly every year~\cite{kristan15vot,kristan2018sixth}.
Nevertheless, a particular class of every-day objects remains challenging even for state-of-the-art methods, namely, rigid flat double-sided objects like cards, books, smartphones, magazines,
coins\footnote{Hence the problem name.}, tools like knives, hand saws, sport equipment like table tennis rackets, paddles etc.
Such objects often rotate fast producing unique challenges for trackers like fast incident light and
aspect ratio change and rotational motion blur.

In this paper, we introduce an annotated \emph{coin-tracking dataset}\footnote{Available at \url{http://cmp.felk.cvut.cz/coin-tracking}.}, CTR dataset in short, containing video sequences of coin-like objects.
We then show that the proposed dataset is fundamentally different from the standard ones \cite{kristan2018sixth,wu15otb}.
Finally, we propose a baseline coin-tracking method, called \method{CTR-Base}, that outperforms classical state-of-the-art trackers in experiments on the CTR dataset.

\section{Coin-Tracking Dataset}
\label{sec:coin-track-datas}
We define coin-tracking as tracking of rigid, approximately planar objects in video sequences.
This means that at any time only one of the two sides - \emph{obverse} (front) and \emph{reverse} (back) - is visible.
Unlike general objects, the rigidity and planarity of the coin-like objects means that the boundary between their two sides is always visible, except for occlusions by another object and position partially outside of the camera field of view.
In this settings, the currently invisible side is fully occluded by the visible side and the visible side does not occlude itself at all.
The state of a coin-like object is thus fully characterized by a visible side identification and a homography transformation to a canonical frame together with a possible partial occlusion mask.

However, because the objects in the CTR dataset are often symmetric, reflecting the real world coin-like object properties, the homography transformation might not be uniquely identifiable and thus we characterize the object state by a segmentation mask instead.
Notice that unlike in standard general tracking sequences, where the exact extend of the tracked object is often not well defined due to the ambiguity of the initialization bounding box or segmentation, there is an unambiguous correspondence between a segmentation mask and a physical object in the case of coin-tracking.

Recent video object segmentation datasets~\cite{perazzi2016davis,xu2018youtube} represent the object pose by segmentation as well, nevertheless, they contain mostly outdoor sequences of animals, people and vehicles.
Therefore, there is a significant domain gap between these datasets and the proposed coin-tracking problem.
Other datasets for tracking planar object exist, such as~\cite{liang2018planar,lin2019robust}, but they only contain sequences with single side of the planar object visible.
Moreover, in most cases the objects are fixed and the camera moves around them.
This induces both different dynamics and appearance changes in the sequences as discussed in section~\ref{sec:comp-with-other}.

The are multiple levels of tracking of coin-like objects. In the simplest form, the tracker is initialized by a template of each side of the object and the object pose on the first frame of the sequence.
One could also initialize the tracker on the first side only and require it to discover the reverse side without supervision.
Moreover, a full 6D pose output (rotation and translation) together with a complete object surface reconstruction (including even the initially occluded parts of the object) could be required for sequences with known camera calibration.

\begin{figure}
  \centering
  \def\svgwidth{\textwidth}
  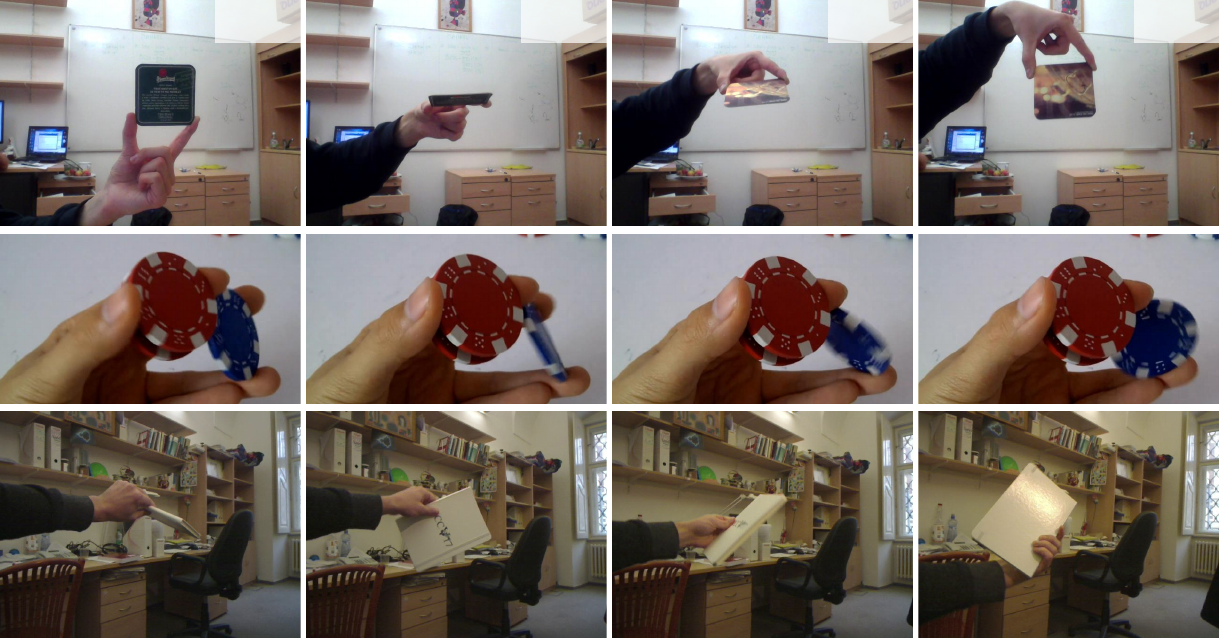
  \caption{Examples from the coin-tracking dataset (frame number in the top-right corner).  Notice the effects of the out-of-plane rotation -- fast illumination change, blur and significant aspect ratio change of the objects.}
  \label{fig:ctr-examples}
\end{figure}

The introduced CTR dataset contains 17 video sequences of coin-like objects, with total of 9257 frames and segmentation ground truth masks on every fifth frame.
See Fig.~\ref{fig:ctr-examples} for examples of the sequences in the CTR dataset.
\subsection{A Comparison with Other Datasets}
\label{sec:comp-with-other}
The main motivation for introducing a new tracking dataset is its difference from the currently available tracking sequences.  In this section we show some of the novel aspects of the proposed dataset.

The planar object tracking datasets~\cite{lin2019robust,liang2018planar} are the closest to the CTR dataset, but they only contain a single sided view of the object; the viewing angle range is limited.
In most of the sequences  the tracked object is fixed to the background behind it,~e.g.~a poster fixed on a wall and the object motion in the sequence is induced by the camera motion only.
On the contrary, the camera is static or close to static in many of the CTR sequences and it is the object that causes the motion.
This difference is important since the two situations introduce different challenges to the visual tracking task.

When a planar object is fixed and a camera moves around it, the perceived out-of-plane rotation is relatively slow as the camera needs to move along a long arc in order to change the viewing angle significantly.
On the other hand, when the main part of the perceived motion of the object in the sequence is caused by the physical motion of the object itself, as it is the case in the proposed sequences, the object out-of-plane rotation happens faster as it is physically easy to rotate coin-like objects.

Most state-of-the-art trackers,~e.g.~the winners of the VOT2018 tracking challenge~\cite{kristan2018sixth} -- MFT~\cite{bai2018multi} and UPDT~\cite{bhat2018unveiling}, represent the object pose as axis-aligned or rotated bounding box, while the aspect ratio change modeling is not common.
Later in this section, we show that both the range and the speed of aspect ratio change in the CTR sequences is higher than in the VOT~\cite{kristan2016vot} and OTB~\cite{wu15otb} tracking datasets.
Besides causing significant aspect ratio changes, the 3D rotation of the coin-like objects often induces fast changes of illumination as the object plane normal direction relative to the light sources changes rapidly.
Apart from these differences, the objects in the CTR dataset are also less textured than the ones appearing in standard visual tracking datasets as discussed in the next section.

\begin{figure}[t]
  \centering
  \includegraphics[width=0.5\textwidth]{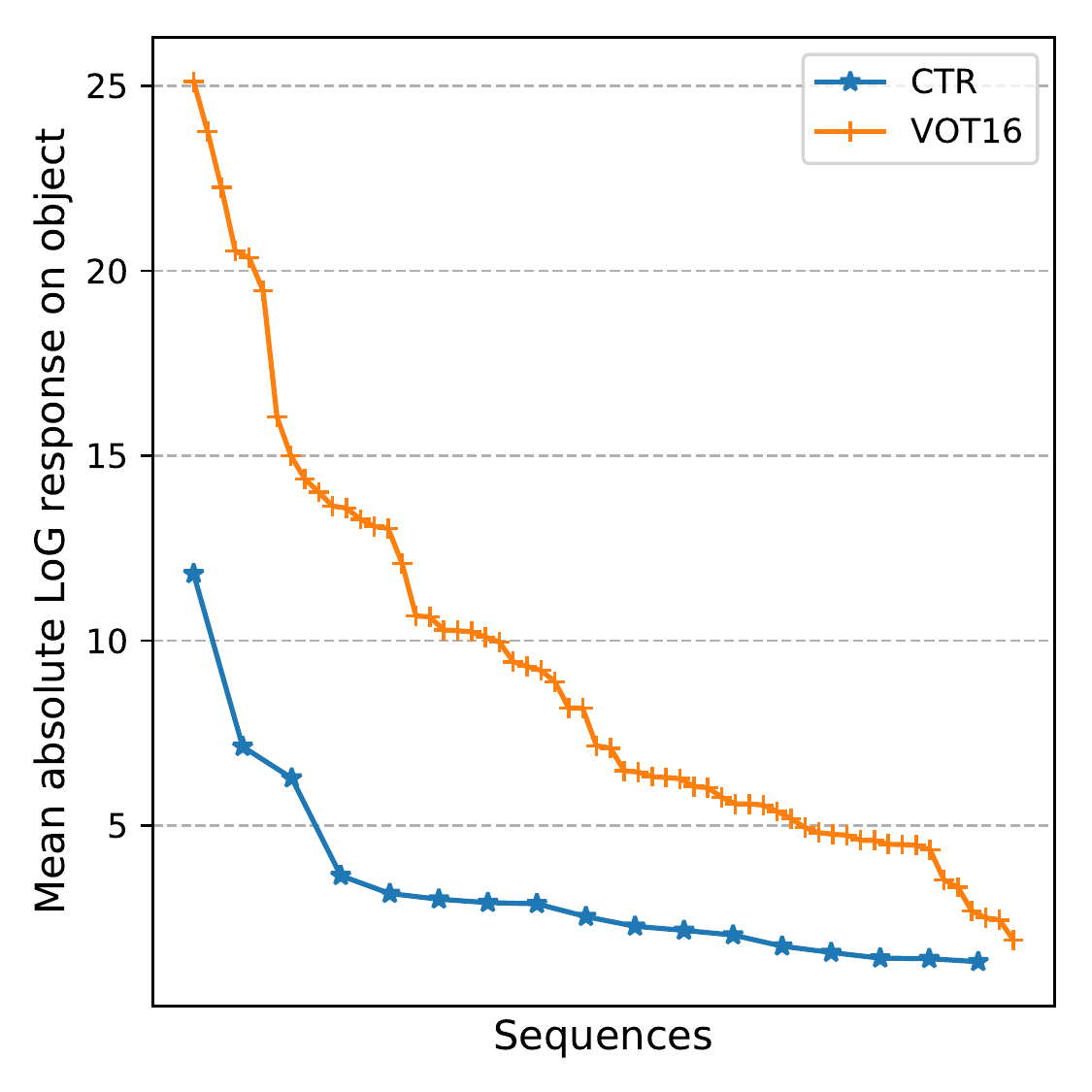}
  \caption{Comparison of object ``textureness'' in the proposed CTR and VOT 2016 datasets, measured by the absolute value of Laplacian of Gaussian \(\sigma = 0.8\) averaged over the tracked object pixels.}
  \label{fig:textureness}
\end{figure}

\noindent \textbf{Textureness.}
As a measure of object textureness, we computed the Laplacian of Gaussian (LoG) responses and averaged their absolute values over the object pixels and all frames.
Fig.~\ref{fig:textureness} shows that the typical object textureness in the CTR dataset is significantly lower than on the VOT 2016 dataset~\cite{kristan2016vot}.
The lack of texture prevents tracking to be implemented by classical methods for homography estimation based on key-point correspondences.

\noindent \textbf{Aspect ratio change.}
One of the unique properties of the coin-tracking dataset is the presence of strong changes in object aspect ratios, not usually encountered in the standard visual tracking datasets as shown in the following two experiments.
In order to compute the aspect ratio statistics, we first compute minimal (rotated) rectangle bounding the ground truth segmentation mask on each frame.
The aspect ratio~\eqref{eq:aspect-ratio} of the resulting rectangle with sides \(a, b\) is defined as
\begin{equation}
  \label{eq:aspect-ratio}
  r(a, b) = \max\left(\frac{a}{b}, \frac{b}{a}\right)
\end{equation}

We define the relative change in aspect ratios of two rectangles \(A, B\) with sides \(a_1, a_2\) and \(b_1, b_2\), respectively, as~\eqref{eq:aspect-ratio-change}
\begin{equation}
  \label{eq:aspect-ratio-change}
  \Delta r(A, B) = \max\left(\frac{r(a_1, a_2)}{r(b_1, b_2)}, \frac{r(b_1, b_2)}{r(a_1, a_2)}\right)
\end{equation}
The maximum of the two ratios is chosen because only the magnitude of the aspect ratio change matters.

\begin{figure}[t]
  \centering
  \begin{subfigure}[b]{0.45\textwidth}
  \includegraphics[width=\linewidth]{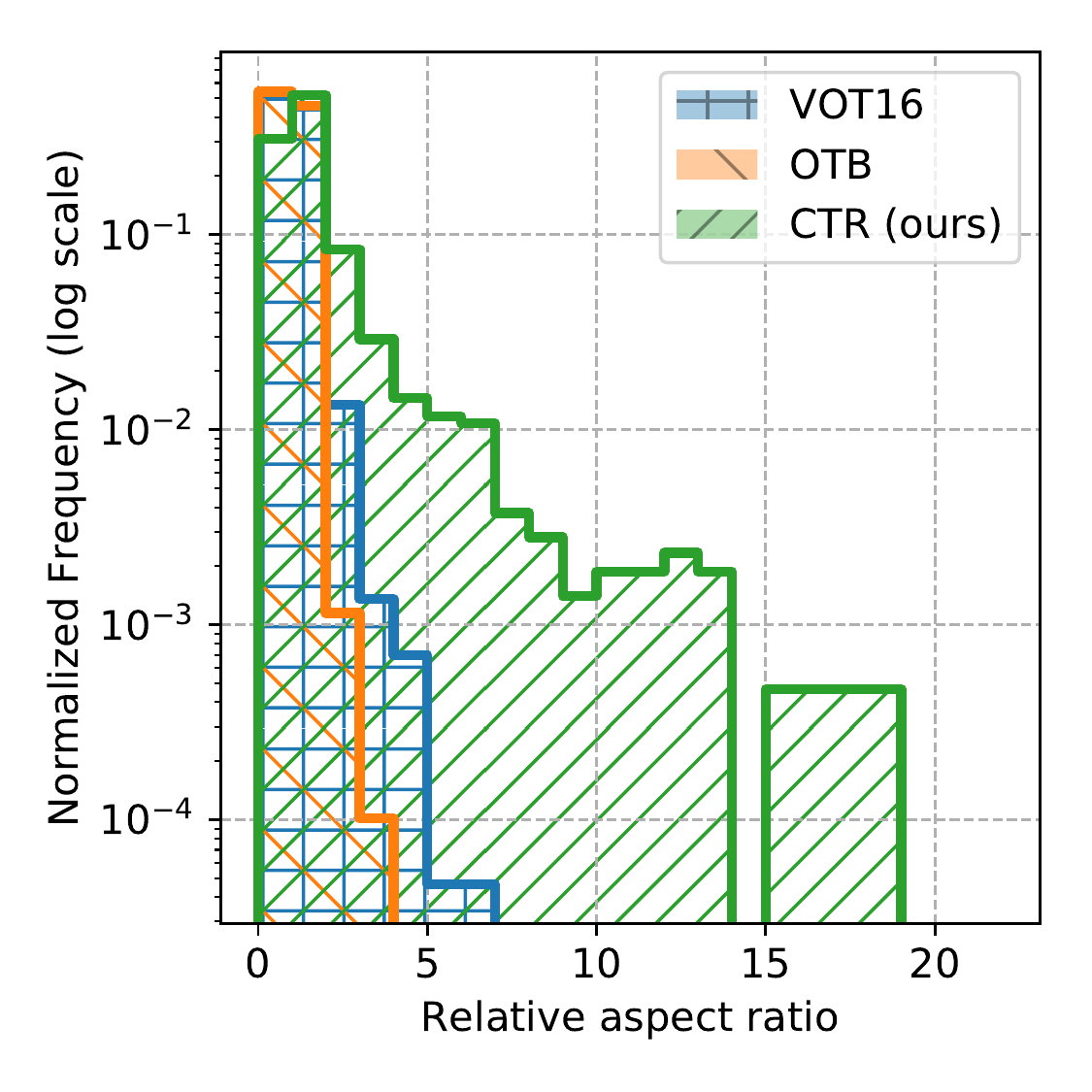}
  \caption{relative to the first frame.}
  \label{fig:aspect-ratio-first}
  \end{subfigure}
  \begin{subfigure}[b]{0.45\textwidth}
  \includegraphics[width=\linewidth]{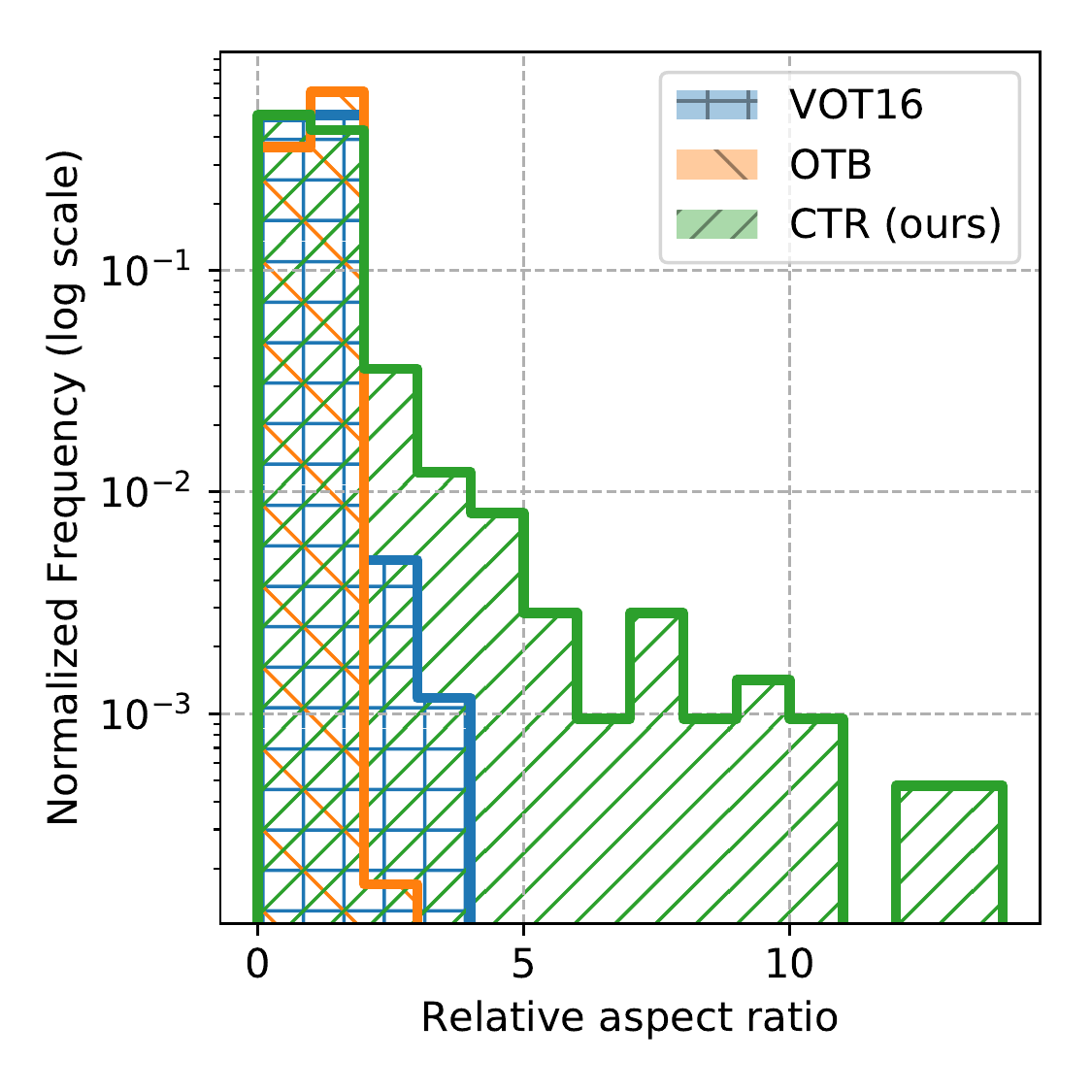}
  \caption{relative to frame t-5.}
  \label{fig:aspect-ratio-last}
  \end{subfigure}
  \caption{Histogram of aspect ratio changes}
\end{figure}

\paragraph{Aspect ratio change relative to the first frame.}
We have computed aspect ratio changes \(\Delta r(R_1, R_t)\) between the bounding rectangle on the first frame and each of the other annotated frames in the sequence.
We then represent each tested dataset (VOT2016, OTB, CTR) by a histogram of these aspect ratio changes in all the dataset sequences as shown in Fig.~\ref{fig:aspect-ratio-first}.
Notice that although the VOT2016 and OTB datasets are not restricted to rigid objects,~i.e. their segmentation masks can change shape arbitrarily during the sequences, the CTR dataset contains significantly bigger changes in the aspect ratios.

\paragraph{Aspect ratio change speed.}
In the proposed CTR dataset, the change in object aspect ratio is also faster than in the other compared datasets as shown in Fig.~\ref{fig:aspect-ratio-last}.
Instead of computing the aspect ratio change with respect to the first frame, the change is computed relative to the previous frame.
Notice that because the CTR dataset does not contain ground truth segmentation masks on every frame, but only on every fifth, we measure \(\Delta r(R_{t-5}, R_t)\) on all three datasets.
\subsection{Evaluation Metric}
We address the simplest form of the coin-tracking task, in which the tracker is initialized by an image of the front side of the tracked object on the first frame and an image of the back side later in the sequence, together with the respective ground truth segmentation masks.

We use \emph{intersection over union} (IoU) as the evaluation metric -- it is the standard metric for evaluating both segmentation and bounding box quality.
In order to deal with frames with empty ground truth segmentation, i.e.~with the object fully occluded or fully outside of the view, we augment the scoring function such that these frames do not contribute into the per-sequence total as proposed in~\cite{kristan2018sixth}.

\section{The Baseline Coin-Tracking Method}
Standard trackers represent the object by a bounding box and are thus unable to capture the perspective transformations common for coin-like objects.
Trackers based on key-point correspondences can estimate homographies, but the low textureness of CTR objects prevents their use.
Convolutional neural networks recently used for video object segmentation, e.g. \cite{caelles2017oneshot,khoreva2017lucid,voigtlaender2017OnAVOS}, classify pixels as object or background taking into account large context thanks to large receptive fields of the neurons in the final layers.
They do not consider the underlying homography transformations, but the segmentations
capture the object extent in the image with high granularity.

Most video object segmentation methods use a deep neural network trained offline for general object segmentation.
The network is then fine-tuned for tracking of a particular object at the initialization.
One of the significant challenges in visual tracking is object appearance change and changes in the background in the video sequence.
Because of this, trackers usually have to perform some kind of \emph{online adaptation} to prevent performance deterioration soon after initialization.
A simple adaptation scheme for video object segmentation has been proposed in \method{OnAVOS}~\cite{voigtlaender2017OnAVOS}, where the pixels classified as object with high confidence are treated as new object appearance examples. Background examples are taken from the parts of the image over a certain distance from the object.
However, the online adaptation requires lengthy fine-tuning of the segmentation neural network on each frame, making the method slow.

An alternative approach has been proposed in \method{fast-vos}~\cite{chen2018blazingly}, where the segmentation is done by k-nearest neighbor search in an embedding space learned offline by a CNN.
Instead of fine-tuning the embedding network on the first frame or later during online adaptation, the \method{fast-vos} method inserts dense embeddings into a k-NN classifier index.
This makes the adaptation to a particular object faster and easier to interpret, compared to the network fine-tuning methods.
The online adaptation proposed in~\cite{chen2018blazingly} is similar to the original method in~\cite{voigtlaender2017OnAVOS}, selecting high confidence pixels -- all of their \(k=5\) neighbors agree with the label -- for the model update.

With all this in mind, we propose a baseline tracking method \mbox{\method{CTR-Base}}, which is based on the tracking-by-segmentation  \method{fast-vos}~\cite{chen2018blazingly} method.
After an input frame is segmented using the k-NN classifier, we explicitly model the object pose and possibly perform online adaptation.

\subsection{Object Pose Estimation}
We have performed experiments with the adaptation scheme of \method{fast-vos} but it did not work well on the coin-tracking sequences.
The adaptation has quickly drifted and led to a complete failure of the tracker,
either segmenting almost all of the background as the object or vice versa.
Our experiments with distance-threshold based background adaptation as in~\cite{voigtlaender2017OnAVOS} as well as experiments with other heuristics based on analysis of the connected components and other properties of the segmentation mask were not successful either.
We hypothesize that one of the reasons that those adaptation techniques work reasonably well on the DAVIS dataset, but fail on the coin-tracking task, might be the length of the sequences.
The mean number of frames in the DAVIS 2017 sequences is only 69.7~\cite{pont-tuset17davis} while the mean number of frames in the coin-tracking sequence in the CTR dataset is 544, with several sequences as long as 1000 frames.
The robustness of the online adaptation scheme is crucial on sequences of such length.
\begin{figure}
  \centering
  \def\svgwidth{\textwidth}
%% Accompanies image file 'score.pdf' (pdf, eps, ps)
\begingroup%
  \makeatletter%
  \providecommand\color[2][]{%
    \errmessage{(Inkscape) Color is used for the text in Inkscape, but the package 'color.sty' is not loaded}%
    \renewcommand\color[2][]{}%
  }%
  \providecommand\transparent[1]{%
    \errmessage{(Inkscape) Transparency is used (non-zero) for the text in Inkscape, but the package 'transparent.sty' is not loaded}%
    \renewcommand\transparent[1]{}%
  }%
  \providecommand\rotatebox[2]{#2}%
  \newcommand*\fsize{\dimexpr\f@size pt\relax}%
  \newcommand*\lineheight[1]{\fontsize{\fsize}{#1\fsize}\selectfont}%
  \ifx\svgwidth\undefined%
    \setlength{\unitlength}{314.75055845bp}%
    \ifx\svgscale\undefined%
      \relax%
    \else%
      \setlength{\unitlength}{\unitlength * \real{\svgscale}}%
    \fi%
  \else%
    \setlength{\unitlength}{\svgwidth}%
  \fi%
  \global\let\svgwidth\undefined%
  \global\let\svgscale\undefined%
  \makeatother%
  \begin{picture}(1,0.42036842)%
    \lineheight{1}%
    \setlength\tabcolsep{0pt}%
    \put(0,0){\includegraphics[width=\unitlength,page=1]{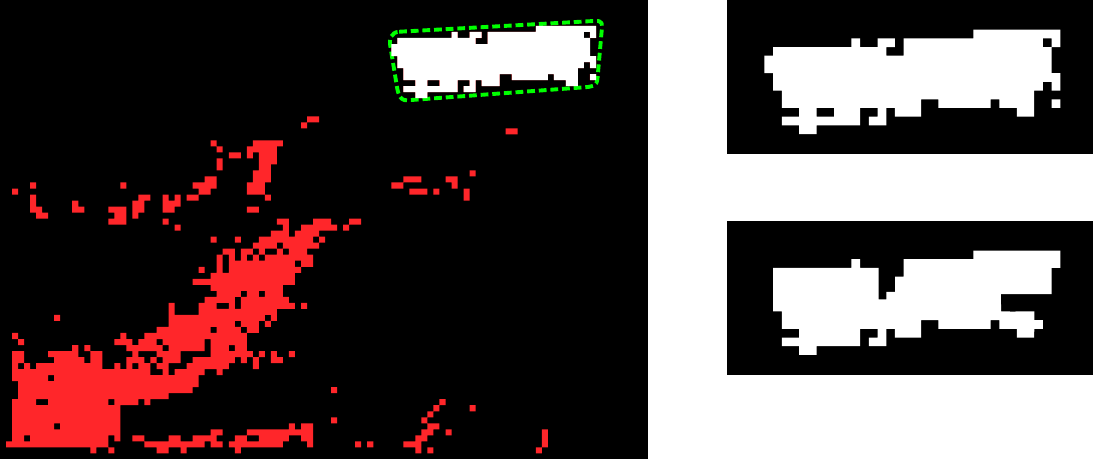}}%
    \put(0.85509952,0.23456445){\color[rgb]{0,0,0}\makebox(0,0)[lt]{\lineheight{0}\smash{\begin{tabular}[t]{l}\#t\end{tabular}}}}%
    \put(0.85509952,0.0312288){\color[rgb]{0,0,0}\makebox(0,0)[lt]{\lineheight{0}\smash{\begin{tabular}[t]{l}\#t-1\end{tabular}}}}%
    \put(0,0){\includegraphics[width=\unitlength,page=2]{score.pdf}}%
  \end{picture}%
\endgroup%

  \caption{Homography score computation.  Left: the segmentation mask split into pixels inside (white) the object pose hypothesis (dashed green) and the rest (red).  Right: Object visibility mask for the current and the last frames.}
  \label{fig:score}
\end{figure}

In order to address the online adaptation in coin-tracking more robustly,
we explicitly model the object pose using the homography to the ground-truth canonical frame.
Both the object and the background pixel online adaptation is controlled by
the agreement between the segmentation output by the k-NN classifier and the estimated pose model.

\subsubsection{Objective Function.}
In each video frame, we search for the homography \( \mathbf{H}_{*\rightarrow t} \) mapping the object on a ground truth frame into the current one, optimizing the objective function \(s\), Eq.~\ref{eq:score}, composed of four parts computed as follows.
First, we map the segmentation mask from the ground truth frame into the current frame using the homography.
This splits the segmentation mask in the current frame into two parts, one inside and the other one outside of the hypothesized object contour as shown in Fig.~\ref{fig:score}.
The \( s_{obj} \) part of the score function is set to the fraction of the segmentation mask located inside the contour, indicating the fraction of the segmentation explained by the object.
This part of the score function penalizes segmentation outside of the object with the pose given by \( \mathbf{H}_{*\rightarrow t} \).

The \( s_{cover} \) part of the score function \(s\) is the fraction of the pixels inside the hypothesized object contour being classified as the object.
This part penalizes homographies mapping the object contour such that it is not well covered by the segmentation.
Notice, however, that in the case of partial occlusion by other object, the segmentation should not cover the whole object.
Since the occlusion mask is changing relatively slowly in CTR sequences,
the \( s_{occl} \) component of the score function \(s\) is the IoU overlap of the current and last visibility mask, which is transformed to the current frame by \( \mathbf{H}_{t-1\rightarrow  t} = \mathbf{H}_{*\rightarrow t}\mathbf{H}_{*\rightarrow t-1}^{-1} \).
This prefers homographies with a small occlusion change with respect to the previous frame.

Finally, the appearance score \(s_{\text{appearance}}\) is the zero-offset
coefficient of the zero-normalized cross-correlation (ZNCC) score
\begin{equation}
  \label{eq:apparance-score}
  s_{\text{appearance}} = \frac{1}{2} + \frac{\sum\limits_{x,y \in O}(I_t(x, y) - \mu(I_t))(I_{*}(x, y) - \mu(I_{*}))}{2 \sqrt{\sum\limits_{x,y \in O}(I_t(x,y)-\mu(I_t))^2 \sum\limits_{x,y \in O}(I_*(x,y)-\mu(I_*))^2}}
\end{equation}
of the object image in the current frame and the template from the ground-truth frame, where \( I_t(x, y) \) and \( I_{*}(x,y) \) are the image values at coordinates \( [x, y] \) in the current frame and the ground truth frame projected using the homography \( \mathbf{H}_{*\rightarrow t} \) respectively and
\begin{equation}
  \label{eq:object-mean}
  \mu(I) = \frac{1}{\left|O\right|}\sum\limits_{x, y \in O} I(x, y)
\end{equation}
with \( O \) being the set of points segmented as object in both the ground truth and the current frame.
The rationale behind introducing the appearance score is that it helps distinguishing a correct homography in case of objects with symmetric shape or partial occlusions.
The final score, Eq.~\ref{eq:score}, of the homography is the product of these four components giving a number in 0-1 range:
\begin{equation}
    \label{eq:score}
    s = s_{obj} \cdot s_{cover} \cdot s_{occl} \cdot s_{appearance}
\end{equation}
Notice that compared to summing the score components, taking their product highlights drops in any of the score components and thus it is preferable for making our adaptation method conservative.

\subsubsection{Optimization.}
Since the cost function described above is not differentiable, we use a probabilistic optimization procedure based on simulated annealing for finding \( \mathbf{H}_{*\rightarrow t} \) for each frame.
The optimization is initialized using either the homography found in the previous frame
or using optical flow from the previous frame, in which case we uniformly sample 4 points from inside the object and transform them by the flow field to get 4 correspondences necessary for estimating the inter-frame homography.
This is repeated 50 times and the \( \mathbf{H}_{*\rightarrow t} \) maximizing the score function is chosen as the initialization of the following iterative optimization procedure.

In each step of the optimization a random homography matrix is sampled by randomly perturbing 4 control points at the corners of the object bounding box and computing the homography from the resulting 4 correspondences.
Next, the homography score \( s \) is computed and compared to the current best score, \( s^{*} \).  The \( \mathbf{H}_{*\rightarrow t} \) hypothesis is accepted as the current estimate of the optimum with probability
\begin{equation}
  \label{eq:p-accept}
  p(s, s^{*}, T) =
  \begin{cases}
    1                        & \text{if}~s > s^{*}, \\
    e^{-\frac{s^{*} - s}{T}} & \text{otherwise,}
  \end{cases}
\end{equation}
where the \( T \) is decreasing in each iteration, allowing jumps from local minima but with decreasing probability during the optimization procedure.
We also decrease the control point perturbation \(\sigma\) in each of the 350 iterations.

Depending on the ratio of pixels being classified as belonging to the obverse or the reverse side of the object, the optimization procedure is run against the respective ground truth frame.
Finally, when the score of the best found homography is low, the tracker switches into a \emph{lost} state and stays in it until a successful re-detection of the object.

The re-detection procedure is the same as the optimization described above, except for spending more time (400 iterations) sampling for the initialization pose and not using the information from the previous frame. The previous visibility mask used in computation of \( s_{occl} \) is replaced by the full object mask.
\subsection{Online Adaptation}
The proposed homography optimization procedure reduces the overall speed of the tracker, but we have observed that it finds a good solution reliably, unless the segmentation is grossly incorrect, enabling us to use online-adaptation on the long sequences in the CTR dataset.
In particular, no online adaptation is attempted when the tracker is in the \emph{lost} state, reducing the probability of making incorrect adaptation.

If the tracker is in the \emph{tracking} state, new background and object embedding examples are added into the segmentation k-NN classifier.
To stay on the safe side, only the pixels that are far from the object boundary and were incorrectly classified (with respect to the hypothesized object pose) are used as new background examples.
Moreover, these pixels must not be connected to the object by the segmentation mask, otherwise they are not used for adaptation even if they are very far from the image.

For the new object examples, we select the pixels classified as background by the segmentation k-NN classifier that are not connected to the object edges, in other words only closed `holes` in the object segmentation are adapted.

Altogether, the proposed online adaptation technique allows for conservative online adaptation, not making severe mistakes that would lead to complete failure of the tracker, as shown in the experiments in section~\ref{sec:results-conf-fram}.

\subsection{Implementation details}
We use a DeepLabv3+~\cite{chen2018deeplab} segmentation head on top of MobileNetv1~\cite{howard2017mobilenets} backbone architecture.
The MobileNet backbone was pretrained\footnote{Code and weights available at \url{https://github.com/tensorflow/models/}} on ImageNet~\cite{deng09imagenet}, then trained for semantic segmentation on PASCAL VOC 2012~\cite{everingham2012pascal} enriched by the \emph{trainaug} augmentations by~\cite{bharath2011semantic}.
We have used the Adam~\cite{kingma2014adam} optimizer with batch size 5 and initial learning rate of \(7 \times 10^{-4}\) decaying to \(10^{-6}\) according to the \emph{poly} schedule with decay power 0.9 for 53000 iterations.
Finally, using the augmented triplet loss proposed by~\cite{chen2018blazingly}, we have fine-tuned the network for 492000 iterations on the YouTubeVOS dataset~\cite{xu2018youtube} to output dense 128-dimensional embeddings useful for segmentation by k-NN classifier.
Given an \(H \times W\) image, the network produces a per-pixel 128-D embeddings with output stride 4 (resolution \(\frac{H}{4} \times \frac{W}{4}\)).
We use FAISS~\cite{johnson2019billion} library\footnote{Available at \url{https://github.com/facebookresearch/faiss}} with a flat L2 index for speeding up the nearest neighbor searches used in the segmentation.
For the optical flow computation, we use ContinualFlow~\cite{neoral2018continual}.

The method runs at around 7 seconds per frame at \(1280 \times 720\) resolution with the majority of time spent optimizing the pose.  The runtime drops without losing much performance when the pose optimization is done on lower resolution.

\section{Experiments}
In this section we show that the proposed \method{CTR-Base} method outperforms general state-of-the-art trackers on the CTR dataset and retains good performance on the \mbox{POT-210}~\cite{liang2018planar} dataset.
Then we demonstrate that the homography-based pose modeling prevents the \method{CTR-Base} tracker from making fatal mistakes.
\subsection{Baseline Experiment}
In the standard visual tracking formulation, the tracker is initialized by the ground truth object pose, which can be represented by  axis-aligned bounding box, rotated bounding box or segmentation mask~\cite{kristan2018sixth,perazzi2016davis,xu2018youtube}.
This means that standard state-of-the-art trackers cannot be directly evaluated on the coin-tracking task in which the tracker is initialized on one frame from each side of the object.
On the other hand, the coin-tracking task can be viewed as a long-term tracking on single side, enabling us to evaluate state-of-the-art long term trackers \method{MBMD}~\cite{zhang2018learning} and \method{DaSiam\_LT}~\cite{zhu2018distractor} -- the winners of the VOT 2018~\cite{kristan2018sixth} long-term tracking challenge on the CTR dataset.
Moreover, the VOT long-term tracking challenge requires a tracker confidence output on each frame, which allows us to run each tracker two times - once initialized from the obverse and once from the reverse side, merging the results by picking the one with higher tracker confidence.
We have represented the axis-aligned bounding box outputs of the long-term trackers as segmentation masks and evaluated using the IoU metric.
The results are shown in Tab.~\ref{tab:eval}.

The proposed \method{CTR-Base} method significantly outperforms both state-of-the-art bounding box trackers and a bounding box oracle, which outputs the bounding boxes of the ground truth segmentation masks.
Computing IoU from the bounding boxes might not seem fair, but the performance gap demonstrates the need of representing the tracked object by segmentation, even with relatively compact objects present in the CTR dataset.
\begin{table}
  \centering
  \begin{tabular}{lrrr|r}
    sequence             & \method{MBMD} & \method{DaSiam\_LT} & bbox oracle & \method{CTR-Base} (ours) \\
    \hline
    beermat              & 0.70 & 0.18       & 0.78        & \textbf{0.83}           \\
    card1                & 0.72 & 0.71       & 0.73        & \textbf{0.79}           \\
    card2                & 0.71 & 0.68       & 0.79        & \textbf{0.93}           \\
    coin1                & 0.60 & 0.62       & 0.71        & \textbf{0.80}           \\
    coin3                & 0.32 & \textbf{0.46}       & 0.63        & 0.38           \\
    coin4                & 0.33 & 0.41       & 0.56        & \textbf{0.65}           \\
    husa                 & 0.35 & 0.40       & 0.51        & \textbf{0.73}           \\
    iccv\_bg\_handheld   & 0.27 & 0.31       & 0.54        & \textbf{0.33}           \\
    iccv\_handheld       & 0.32 & 0.39       & 0.55        & \textbf{0.50}           \\
    iccv\_simple\_static & 0.37 & 0.31       & 0.51        & \textbf{0.65}           \\
    iccv\_static         & 0.34 & 0.40       & 0.55        & \textbf{0.67}           \\
    pingpong1            & \textbf{0.42} & 0.38       & 0.64        & 0.33           \\
    plain                & 0.44 & 0.50       & 0.60        & \textbf{0.74}           \\
    statnice             & 0.53 & 0.57       & 0.67        & \textbf{0.87}           \\
    tatra                & 0.47 & 0.54       & 0.66        & \textbf{0.86}           \\
    tea\_diff\_2         & 0.54 & 0.57       & 0.61        & \textbf{0.87}           \\
    tea\_same            & 0.53 & 0.52       & 0.63        & \textbf{0.85}           \\
    \hline
    Mean over all frames & 0.47 & 0.44       & 0.63        & \textbf{0.70}           \\
  \end{tabular}

  \caption{The evaluation of the IoU overlap metric on the proposed CTR dataset.  Notice that the \method{CTR-Base} method outperforms both state-of-the-art long-term trackers and the bounding box oracle.}
  \label{tab:eval}
\end{table}

In order to further test the \method{CTR-Base} method, we evaluated it on the \mbox{POT-210}~\cite{liang2018planar} dataset, converting the ground -- object corners -- to segmentation (not modeling occlusions).
The mean IoU (mIoU) is \(0.81\), showing that our method generalizes to \mbox{POT-210} well.
The best results were achieved on the \emph{out-of-view} and the \emph{perspective distortion} subsets of~\cite{liang2018planar} with mIoU \(0.89\) and \(0.88\) respectively, while the worst on the \emph{motion blur} subset with mIoU of~\(0.71\).

\subsection{Results on confident frames}
\label{sec:results-conf-fram}
The mean IoU score computed only on the frames where the \method{CTR-Base} method is in the \emph{tracking} state,~i.e.~online adaptation is allowed, improves from 0.70 to 0.88.
This shows that the proposed tracker can correctly detect its own failures and only adapt when tracking reliably.
Overall the tracker spends 47\% of the frames in the \emph{tracking} state as shown in Tab.~\ref{tab:confident-results}.
\begin{table}
  \centering
  \begin{tabular}{c|ccccccccccccccccc|c}
\rotatebox{90}{sequence} & \rotatebox{90}{beermat} & \rotatebox{90}{card1} & \rotatebox{90}{card2} & \rotatebox{90}{coin1} & \rotatebox{90}{coin3} & \rotatebox{90}{coin4} & \rotatebox{90}{husa} & \rotatebox{90}{iccv\_bg\_handheld} & \rotatebox{90}{iccv\_handheld} & \rotatebox{90}{iccv\_simple\_static} & \rotatebox{90}{iccv\_static} & \rotatebox{90}{pingpong1} & \rotatebox{90}{plain} & \rotatebox{90}{statnice} & \rotatebox{90}{tatra} & \rotatebox{90}{tea\_diff\_2} & \rotatebox{90}{tea\_same} & \rotatebox{90}{\textbf{average}}\\
\hline
IoU \(\times 100\) & 89 & 89 & 96 & 82 & 94 & 84 & 87 & 90 & 85 & 85 & 83 & 67 & 88 & 89 & 92 & 92 & 86 & 88\\
frames in \emph{tracking} state \% & 89 & 68 & 93 & 64 & 02 & 21 & 69 & 17 & 15 & 29 & 28 & 17 & 42 & 46 & 34 & 87 & 47 & 47 \\
\hline
  \end{tabular}
  \caption{The IoU score of the \method{CTR-Base} tracker evaluated only on the frames, where it is in the confident \emph{tracking} state and the online adaptation is enabled.  Notice that indeed the tracker is confident on the frames, where it performs well.}
  \label{tab:confident-results}
\end{table}
\section{Conclusion}
We have introduced a novel video analysis problem -- coin tracking -- and presented a novel tracking CTR dataset consisting of 17 sequences of \emph{coin-like} objects and ground truth segmentations.
We have shown its dissimilarity to other tracking datasets.
Besides studying the special properties of coin-like objects, the CTR dataset may benefit both training and
the evaluation of general trackers, including video object segmentation methods, because it contains objects classes different from the ones encountered in the available datasets.  Sequences in CTR are long, making online adaptation more challenging.

We have proposed a baseline \method{CTR-Base} tracking method that enables robust online adaptation through explicit modeling of the tracked object pose and failure detection.
The proposed \method{CTR-Base} method outperforms state-of-the-art long-term trackers on the CTR dataset in terms of the IoU while generalizing well to the \mbox{POT-210} dataset~\cite{liang2018planar}.

Finally, the advanced variants of the coin-tracking task described in section~\ref{sec:coin-track-datas},
like the unsupervised back side discovery or full surface reconstruction, are challenging and open topics left for future research.

\subsubsection{Acknowledgements.}
This work was supported by Toyota Motor Europe HS, by CTU student grant SGS17/185/OHK3/3T/13 and Technology Agency of the Czech Republic project TH0301019.

\bibliographystyle{splncs04}
\bibliography{073-main}

\end{document}